\documentclass{gicam}

\begin{document}
%Please ignore this line
\GIC{1}

\runningheads{L.\ Serafino}{A paper for the
proceedings of GICAM}

\title{No Free Lunch Theorem and Bayesian probability theory: two
sides of the same coin. Some implications for black-box optimization and metaheuristics\footnote{The First version of this paper has been presented at Gulf International Conference On Applied Mathematics,
November 19-21 2013, Kuwait}}

\author{Loris Serafino}

\address{Department of Mathematics,
Australian College of Kuwait.\\
E-mail: l.serafino@ack.edu.kw}

%\cgsn{Publishing Arts Research Council}{98--1846389}

\begin{abstract}

Challenging optimization problems, which elude acceptable solution via conventional calculus methods, arise commonly in different areas of industrial design and practice. Hard optimization problems are those who manifest the following behavior: a) high number of independent input variables; b) very complex or irregular multi-modal fitness; c) computational expensive fitness evaluation. This paper will focus on some theoretical issues that have strong implications for practice. I will stress how an interpretation of the No Free Lunch theorem leads naturally to a general Bayesian optimization framework. The choice of a prior over the space of functions is a critical and inevitable step in every black-box optimization. 
\end{abstract}


\section{Introduction}
One of the most important stages in many areas of engineering and applied sciences is modeling and the use of optimization techniques to increase the quality and performance of products or processes. Generally in literature the term \textit{optimization} is related to (the output of) a  mathematical technique or algorithm used to identify the extreme value of an  arbitrary objective function (\textit{fitness}) through the manipulation of a known set of variables  and subject to a set of constrains.  More technically, a maximization problem with an explicit objective can in general be expressed in the following mathematical form: finding the value

\[
\operatorname*{arg\,max}_{x\in\mathcal{H}}f(x),
\]

\noindent where $x$ is a give vector in a generic multidimensional space $\mathcal{H}$ and $f:\mathcal{H}\rightarrow\mathbb{R}$ is a function of the vector $x$ and $\mathcal{H}\subset\mathbb{R}^{n}$is a (discrete or continuous, but here the focus will be on continuous) subset of the multidimensional real Euclidean space. From now on we will refer to $\mathcal{H}$ as the \emph{search
space}\footnote{Even if there is no explicit mention of constrains here the formulation is nonetheless enough general since they can be incorporated through an appropriate definition of the search space $\mathcal{H}$}. In most real-world  engineering optimization problems, no analytical expression exists for accurately evaluating the response of a candidate solution. Sometimes the fitness consists just in the possibility to observe different sets of pairs of input and output from a computational simulation or an experiment. This is the \textbf{\textit{black-box scenario}} that will be considered here. Further, in what follows we will refer to as \textbf{\textit{hard optimization problems}} those with a fitness that manifest a combination, to a varying degree, of the following elements:

\begin{itemize}
	\item \textbf{High number of independent input variables.} The large number of candidate solutions to an optimization problem makes it computationally very hard to be attacked by evolutionary algorithms because the number of candidate solutions grows exponentially with increasing dimensionality. This fact, which is frequently named \textit{the curse of dimensionality}, is well known by practitioners that have to handle problems with hundreds of variables \cite{hast}.
	\item \textbf{Difficult fitness landscape. }In landscape surface with weak (low) causality, small changes in the candidate solutions often lead to large changes in the objective values, i.e. ruggedness \cite{wis}. A fitness landscape is rugged if there are many local optima concentrated in any constrained region of the space. At the opposite side, even \textit{neutral} landscape where the optimum is a narrow peak in between regions with similar fitness values can be very hard to optimize since no guiding information can be extracted from neutral areas.
	\item \textbf{Computational expensive fitness evaluation.} Simulation can be very time consuming in the order of many minutes or even hours for each single fitness call.
\end{itemize}
In hard black-box optimization many issues are at stake. Some good reviews are \cite{wis,sha,yoe}. In fig 1 are presented some practical hints derived from literature. It's worth noting that increasing sample size as countermeasure for difficult landscapes compete with the need of reducing the number of fitness calls in computational expensive problems. Further, a number of engineering design tasks as well in other context are modeled as multi-objective problems and this makes the optimization process even harder but this case will not be considered here. Here the focus will be mainly on the following issues: the practical meaning of the \textbf{\textit{No Free Lunch theorem}} (NFLT) and on its natural connection with the Bayesian framework. This choice is due to the fact that they both represent a critical link between optimization theory and optimization practice.

\begin{figure}
	\centering
		\includegraphics[width=0.50\textwidth]{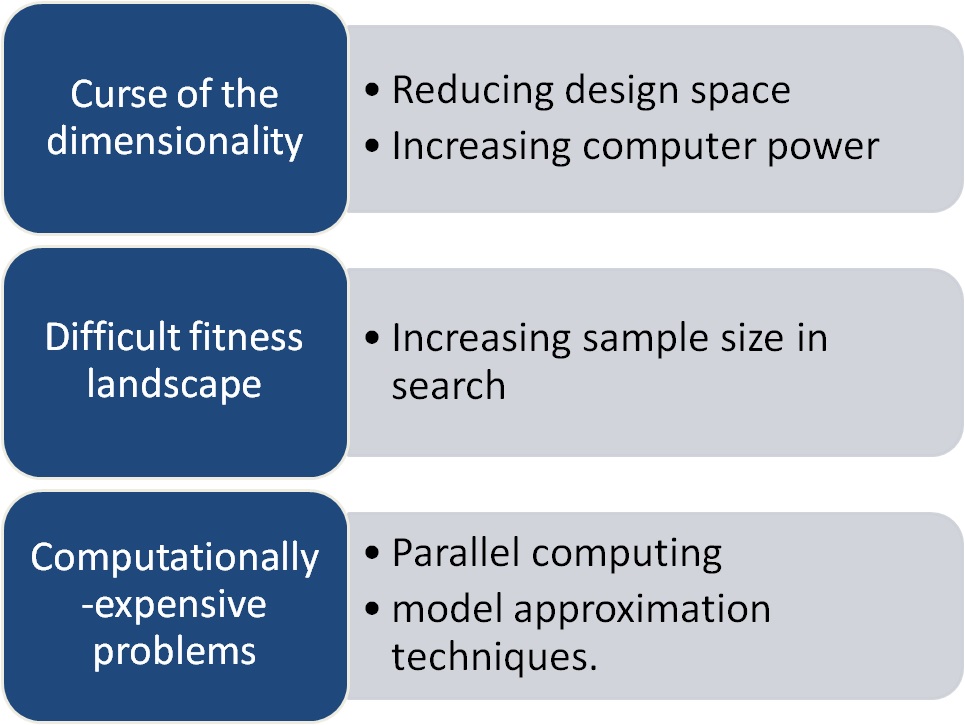}
	\caption{ Some basic strategies for hard optimization.}
	\label{fig:GP}
\end{figure}

\section{Black-box optimization is a blind search, and search is a sampling process}

Black-box optimization (easy or hard) is the reign of metaheuristics. A Metaheuristic describes the way an optimization method decides which part of the search space to explore in the next step. In general every optimization algorithm can be abstractly considered as a sampling process acting on the search space. It starts with a set of values, and then step after step it generates new samples according to some specified mechanism based on current samples and the fitness values: 

	\[\left(x_{1}^{i+1},\dots x_{m}^{i+1}\right)=\text{Alg}\left(x_{1}^{i},\dots, x_{n}^{i}, f\left(x_{1}^{i}\right),\dots, f\left(x_{n}^{i}\right),  \Theta\right)
\]

\noindent where $\Theta$ is a random variable and index $i$ is the iteration counter.

So differences among different optimization methods come from the specific mechanism an algorithm uses to generate and accept new samples and in so doing alternating an exploration (global) phase with a exploitation (local) phase. For example just to mention one class, \textit{nature-inspired} algorithms, like the well known genetic algorithm \cite{mit}, are based on the idea of mimicking some natural phenomena that leads to the maximization of some defined quantity. They can be population-based or not. Other examples are: Particle Swarm, Ant Colonies, Simulated Annealing and many other families \cite{xsy}. However, practice shows that any successful application depends on careful tuning of operators, parameters, and problem-dependent features.

Potentially there are an infinity of possible optimization problems, one for any possible fitness function. At the same time there are an infinity of thinkable optimization methods, one for every possible exploration-exploitation trade-off combination. The choice of the correct metaheuristics for a given class of problem is a crucial theme that leads to take into account the role played by the NFLT in terms of its theoretical and practical relevance.

\begin{figure}
	\centering
		\includegraphics[width=0.80\textwidth]{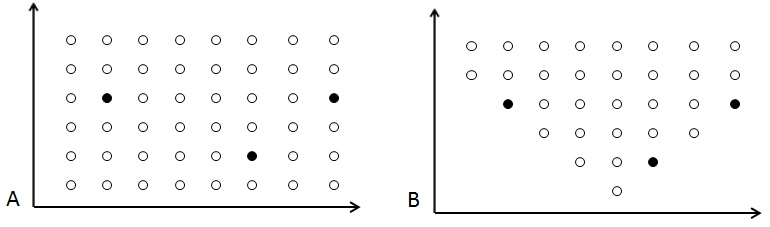}
	\caption{The No Free Lunch Theorem. The information collected so far will not say anything about the values of the function in other regions (case A). For a given subclass of function i.e. convex (Case B), those algorithm that can take advantage of this structure will perform better than others.}
	\label{fig:NFLT}
\end{figure}

\section{The practical meaning of the No Free Lunch Theorem}

We will start with the usual formal statement of the NFLT for optimization \cite{nfl}. Wolpert and Macready's result considered a finite search space ${\cal X}$ and the space of  all the  possible objective functions $ f: {\cal X} \mapsto {\cal Y}$ defined on it called ${\cal F}={\cal Y}^{\cal X}$. They defined with $P(y_k |f,k,Alg)$ the conditional probability of finding a value $y_k\in{\cal Y}$ given a function $f$, after $k$ iteration with algorithm $Alg$. This can be seen a performance measure of the algorithm, its ability to locate a given function value after a given amount of iterations. Under some quite general conditions the theorem states that, for any pair of algorithms $Alg_1$ and $Alg_2$:

$$
 \sum_f P(y_k|f,k,Alg_1) =\sum_f P(y_k |f,k,Alg_2).
 $$

Where the sum is carried out over the set of all the possible function ${\cal F}$. 

According to the most common understanding the NFLT implies that there is no optimization method superior to others for all possible optimization problems. For some function $Alg_1$ will be able to locate the maximum faster than $Alg_2$; for some other function it will be the opposite. Averaging over the whole space ${\cal F}$ the performance will be the same. Equivalently it is possible to say that over ${\cal F}$ no algorithm will perform better than pure random search. Wolpert and Macready adopted a probabilistic framework. Their result holds if we assume a uniform distribution over ${\cal F}$, i.e. any functional form is uniformly admissible. To understand the practical implication of this theorem for black-box optimization problem let's re-state it adopting a different perspective. Substantially the theorem states that

\vspace{0.6cm}
\noindent\textbf{NFLT - \textit{No induction} form} \textit{With no prior knowledge about the function $f: {\cal X} \mapsto {\cal Y}$, in a situation where any functional form is uniformly admissible, the information provided by the value of the function in some points of the domain will not say anything about the value of the function in other regions of the its domain.}
\vspace{0.6cm}

This interpretation of the NFLT is pictured in fig. 2. So in this scenario the information collected with the data sample is not helpful in guiding the search in which direction is better to explore next. In this sense, averaging over the set of all the possible functions, every algorithm performs the same. Of course every function has its own structure, the problem if is when prior knowledge about the functional form is not available because no rationale can guide an optimization strategy i.e. to decide which optimization method to use, which set of parameters, and so on. 

The lesson that has to be learned from NFLT is in the implications for a rational optimization strategy able to tackle hard optimization problems. It is clear now that for the practitioner the correct question is not \textit{which algorithm I have to use} but \textit{what is the geometry of the problem fitness}: the optimization problem becomes an inferential problem as it will be clear in the discussion below. Knowing the structure of the fitness landscape makes (theoretically) possible to properly tune an algorithm in terms of a trade-off between local search and global search. It is also true that general results about which class of algorithms is better suited to which kind of problems class is still far to come even if many studies are going on. As we will see below, Bayesian probability theory will appear to be the natural foundational framework for  metaheuristics. 

In the NFLT scenario, where nothing is known in literature about the structure of the problem at hand, practitioners tend to decide the optimization method according their background knowledge, practical availability of code, simulation software and so on. In a different scenario where something is known about the function, like a lower bound on the function value or some information about the response landscape, this information must be used to tailor the algorithm and the optimization strategy accordingly. Knowledge of the objective function structure is a key to adjust effective algorithms in terms of a better trade-off between exploration and exploitation. Summarizing, from the discussion of the NFLT, black-box optimization involves two main ingredients: input-output data and some prior knowledge. This leads naturally to the next section where the connection with the Bayesian framework will be explored.

\begin{figure}
	\centering
		\includegraphics[width=0.70\textwidth]{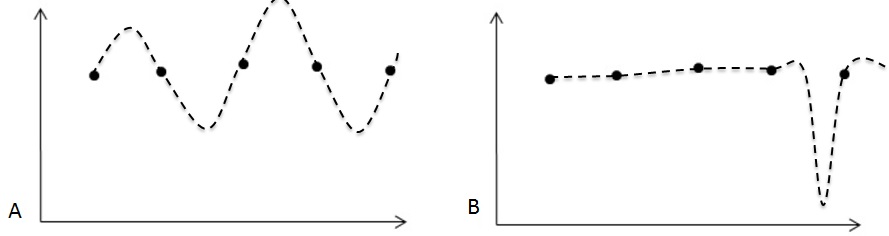}
	\caption{The NFLT is strictly connected to the problem of underdetermination: a sample of data can be described by quite different models (in this case function A and function B).}
	\label{fig:dots}
\end{figure}

\section{NFLT and the Bayesian philosophy: two sides of the same coin}

Our interpretation of the NFLT shows strong connection with the problem of \textit{underdetermination} (fig. 3).  In the black-box context only a finite sample data set concerning the functional response is available: $\mathcal{D}=\left\{\left(x_{1},f\left(x_{1}\right)\right)\dots,\left(x_{n},f\left(x_{n}\right)\right)\right\}$.In this case locating the optimum becomes a pure inferential problem that leads naturally to a Bayesian framework. The rationale of Bayesian probability theory is briefly summarized here in the context of optimization. The problem is to construct a model of the function we need to optimize. In the Bayesian framework there are two ingredients: a data set and a prior. A prior distribution $P(f)$ over the space of functions is combined with the likelihood $P(\mathcal{D}|f)$ to generate a posterior: 
$$P(f|\mathcal{D}) \propto P(\mathcal{D}|f) P(f),$$

\noindent that takes into account the information given by the data set $\mathcal{D}$. In term of modeling a function, if we assume as a prior every possible function (that is equivalent to say that there is no prior) we are exactly in the NFLT case: no useful inferential information can be extracted from the data set. If the prior can be restricted to a proper class of function the inference about the model realizing the data set will be more accurate. As said about the implications of the NFLT, if it is possible to restrict the problem to a given sub-class of function, a proper optimization algorithm choice can produce good results. \textbf{For an optimization method, given $\mathcal{D}$, to have indications about where promising areas for the optimum can be located is equivalent to say that a (posterior) refined model of the fitness has been inferred in Bayesian terms}. 

\begin{table}[ht]\small
\begin{tabular}{|l|l|l|l|}
\hline
\hline
\multicolumn{4}{|c|}{}\\
\multicolumn{4}{|c|}{Given a sample data set}\\
\multicolumn{4}{|c|}{$\mathcal{D}=\left\{\left(x_{1},f\left(x_{1}\right)\right)\dots,\left(x_{n},f\left(x_{n}\right)\right)\right\}$}\\
\multicolumn{4}{|c|}{}\\
\hline
\multicolumn{2}{|c|}{\textbf{NFLT terminology: optimum location}} &
\multicolumn{2}{c|}{\textbf{Bayesian terminology: model generation}}\\
\hline
\multicolumn{1}{|c|}{Uniform over ${\cal F}$ } & Non-uniform over ${\cal F}$&
With no prior&
\multicolumn{1}{c|}{With prior }\\
\hline
No $Alg$ better than & Some $Alg$ better than & No valid inference & Inference can be            \\
pure random search   &pure random search      & for the model 		 & made about the model\\
in locating $\operatorname*{arg\,max}f(x)$&in locating $\operatorname*{arg\,max}f(x)$  & realizing $\mathcal{D}$									 & realizing $\mathcal{D}$                 \\ 
 \hline
\end{tabular} 
\caption{ NFLT vs. Bayesian theory}
\end{table}

As it is now clear, \textit{there is a strong connection between the NFLT and the Bayesian framework} at the point that they can be considered two sides of the same coin. This is summarized in the table 1.
The NFLT assumes a uniform distribution over the space of possible functions. In the Bayesian terminology this is equivalent to say that there is no prior knowledge about the fitness: every functional form is admitted. According to the implication of the NFLT, it is essential to restrict the class of possible function (in the NFLT terminology) to a proper subclass in order to tailor an effective optimization strategy. This restricted class of function represents the prior that appears in the Bayesian formula. So the secret of a successful black-box optimization, assuming a \textit{representative} sample $\mathcal{D}$\footnote{For Design of Experiments techniques see \cite{dace}}, lies in the possibility of narrowing the prior. 

\section{\textit{NFLT - Bayesian framework} duality in optimization: some implications}

The discussion above about the interpretation of NFLT in terms of the Bayesian approach (and vice-versa) can be useful in two ways: to use the NFLT for understanding surrogate-based optimization techniques and, on the other side, to use a Bayesian framework for understanding the metaheuristics working logic.
In the context of hard optimization there is a large literature about methods for fitness approximation (also called \textit{meta-models} or \textit{surrogates}) as ways to generate functional models that are computationally more efficient. All of them \textit{assume}, seldom implicitly, a prior in terms of a restricted class of modeling function.  Functional surrogate models can be algebraic representations of the true problem functions. The most popular ones are polynomials, in a method often known in the statistical literature as \textit{response surface methodology}. Several related methods are now commonly used for fitness approximation: the Kriging model, radial-basis-function networks and the support vector machines (for a general reference, see \cite{dace,jin}). After the model has been developed some classical derivative-based or derivative-free methods can be applied \cite{dfo}. 
\emph{Bayesian optimization} is briefly discussed here as a tool that is getting relevance in optimization practice. This technique conceptually combines surrogate estimation with optimum localization (for a good introduction see \cite{bo,alp,hof}). It uses the Bayesian rationale to infer about a starting fitness model. The second step is to infill new points of the search space in those regions where a maximum improvement is expected according to the current best value and the predictive variance (fig. 3). Then the Bayesian procedure is updated. Usually in literature the prior is supposed to be a realization of a \textbf{Gaussian process} that means a distribution over the space of function modeled as a  multivariate Gaussian distribution, so one can think the Gaussian process as defining a distribution over the space of functions:

\[
f(x) \sim \operatorname{GP}(m(x), k(x_i,x_j)).
\]

\noindent meaning that the function $f$ is distributed as a GP with mean function $m(x)$ and
covariance function $k(x_i,x_j)$. A common choice for the covariance function is defined by:  

\[
k(x_i, x_j) = \exp\big(-\tfrac{1}{2}(x_i-x_j)^{T}
\operatorname{diag}(\phi)^{-2}(x_{i}-x_{j})\big), 
\]

\noindent Where $\phi$ represents a family of hyper-parameters that have to be properly estimated.

\begin{figure}
	\centering
		\includegraphics[width=0.47\textwidth]{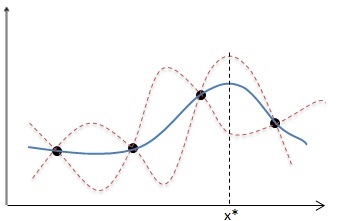}
	\caption{Bayesian optimization. Data points are shown as black dots. The GP modeled distribution has a mean shown by the blue line that fits the data, and a standard deviation represented here by the red dotted lines. Here $x*$ represent the next point to explore since maximum improvement is expected.}
	\label{fig:GP}
\end{figure}

The power of Bayesian optimization philosophy briefly sketch out above lies in its ability to tackle optimization problems with a limited number of fitness calls and this is very important for computational expensive simulations. Again, the key point in the Bayesian rationale is given by the choice of the prior, i.e the non uniform distribution over over ${\cal F}$ in the NFLT terminology. The choice of the prior as a \textit{Gaussian process} implies that the fitness is supposed to be in the class of smooth well-behaving functions and this can be an arbitrary hypothesis. For objective functions with discontinuities or jumps this prior will not be able to provide very good results. Further, the construction of the surrogate model is essentially an exercise of design space exploration exactly in the same way any optimization algorithm result in essence is. If the dimension of the search space is high (more than 100 variables) the ability to sample promising regions becomes weaker and weaker: Bayesian optimization cannot avoid the curse of dimensionality.

From the discussion above it is now clear how a Bayesian framework can be used to shed light on the main metaheuristics operating principles. All metaheuristics \textsl{assume} a model class (the Bayesian prior) of the objective function during the search and combine it with the sampled points at a given iteration to direct the search in the most promising directions. For example \textit{genetic algorithms}, as well as other optimizers, assume (and they work well if there is) \textit{strong causality}: small causes produce small effects. For function with low causality they tend to suffer \cite{wis}. In metaheuristics the prior about the fitness is implicitly define by the search operators used (\textit{crossover}, \textit{mutation}, etc. ) and the set of internal working parameters. Many automated self-adapting strategies able to change the internal parameters values according to partial results obtained during the search process have been studied in recent years \cite{eib}. What these (so called \textit{smart}) optimization algorithms do is to incorporate a more explicit Bayesian operating logic. They generate a posterior model using the sampled points collected at a given iteration and some assumptions about the fitness structure (prior). Iteration after iteration, they adjust the set of internal parameters to better tailor the fitness model and to foster the search in more promising regions (according to the \textit{posterior} model). 

Just for mentioning one, Covariance Matrix Adaptation Evolutionary Strategies (CMA-ES) is one of the most famous in the field of continuous global optimization \cite{hans}. Again, the main idea of this algorithm is to use the information collected during the iterations in terms of sampled points to generate \textsl{on-the-fly} a model of the function to optimize by assuming a local second-order functional structure.
This again shows the strong relationship between optimum location, model determination and the prior knowledge about the fitness geometry over which the optimization process is supposed to be carried out. The internal fitness prior of a metaheuristic defines and constrains the exploration-exploitation trade-off  an so the overall capabilities of the search process. 

Summarizing, the ability of every optimization method is connected to the following points: a) how much the adopted prior model fits the geometry of the problem at hand and b) on the other side, the possibility to access to a \textit{good} sample $\mathcal{D}=\left\{\left(x_{1},f\left(x_{1}\right)\right)\dots,\left(x_{n},f\left(x_{n}\right)\right)\right\}$ of $\mathcal{H}\times\text{Im}(f)$. This can be a problem for computational expensive functions and/or for high-dimensional search spaces. As we have said. the curse of dimensionality affects the ability to generate a valid posterior since with increasing dimensions the needed number of sampled points able to say something usefully in the inferential process increases exponentially.

\section{Conclusions}

Summarizing the discussion above:

\begin{itemize}
\item Black-box optimization $=$ prior fitness knowledge $+$ Sample data set $+$ Algorithm.
\end{itemize}

\begin{itemize}
\item In general, knowledge about the fitness landscape is the key for a better optimization in terms of the possibility to better tune the trade-off between local search (exploration) and global search (exploitation).
\end{itemize}

\begin{itemize}
\item The NFLT and Bayesian theory can be seen as two faces of the same coin. At an abstract level, the Bayesian approach is the natural framework on which to base an effective black-box optimization strategy. 
\end{itemize}

\begin{itemize}
\item The choice of the \textbf{\textit{prior}} is the critical point. The Gaussian process in Bayesian optimization is a valid choice whereas the fitness to be optimized is in the class of smooth functions. 
	\end{itemize}

	\begin{itemize}
\item All metaheuristics more or less implicitly assume a fitness model (prior).They work well if this model matches the geometry of the problem under optimization.
\end{itemize}

	\begin{itemize}
\item The curse of dimensionality affects the ability to search of every optimization method: in  high-dimensional search spaces the number of sampled points needed for the inference process increases dramatically. 
\end{itemize}

\end{document}